%% file: SpaFactor_AnonymousSubmission2027.tex
\documentclass[letterpaper]{article} 

\usepackage{aaai2027}

\usepackage[hyphens]{url} 
\usepackage{graphicx} 
\def\UrlFont{\rm} 
\usepackage{natbib} 
\usepackage{caption} 
\usepackage{booktabs}
\usepackage{amsmath,amssymb}
\usepackage{placeins}
\newcommand{\meanstd}[2]{#1 $\pm$ #2}

\newcommand{\valsd}[2]{#1$_{\scriptscriptstyle\pm \rlap{$\scriptscriptstyle #2$}\phantom{00.00}}$}
\newcommand{\ablationhead}[1]{{\fontsize{8}{9}\selectfont #1}}
\newcommand{\ablationheadleft}[1]{\multicolumn{1}{l}{\ablationhead{#1}}}
\newcommand{\ablationheadfirst}[1]{\multicolumn{1}{@{}l}{\ablationhead{#1}}}

\title{SpaFactor: Lightweight Spatial Context-Aware Gene Program Modeling for Histology-to-Transcriptomics Inference}

\author{
Shiting Ruan\textsuperscript{\rm 1}\equalcontrib,
Xitong Ling\textsuperscript{\rm 1}\equalcontrib,
Qiming He\textsuperscript{\rm 2},
Ziyou Yan\textsuperscript{\rm 1},
Huaitian Yuan\textsuperscript{\rm 1},
Tian Guan\textsuperscript{\rm 1},
Ying Xiao\textsuperscript{\rm 3}\corresponding,
Xu Guan\textsuperscript{\rm 4}\corresponding,
Yonghong He\textsuperscript{\rm 1}\corresponding
}

\affiliations{
\textsuperscript{\rm 1}
Shenzhen International Graduate School, Tsinghua University,
Shenzhen, China
\\
\textsuperscript{\rm 2}
Fuzhou University,
Fuzhou, Fujian, China
\\
\textsuperscript{\rm 3}
Beijing Tsinghua Changgung Hospital,
School of Clinical Medicine, Tsinghua University,
Beijing, China
\\
\textsuperscript{\rm 4}
Chinese Academy of Medical Sciences and Peking Union Medical College,
Beijing, China
}

\begin{document}

\maketitle

\begin{abstract}
Spatial transcriptomics (ST) profiles gene expression within tissue architecture, but its cost and experimental complexity limit routine use. Predicting spatial expression from routinely available hematoxylin and eosin (H\&E) images therefore offers a scalable alternative. However, conventional methods often fit high-dimensional gene outputs as independent targets, overlooking the biological coordination among genes while remaining vulnerable to high-dimensional noise and overfitting. Existing attempts to address this limitation often rely on computationally heavy graph networks or complex auxiliary supervision. We therefore introduce SpaFactor, a lightweight and efficient low-rank morphology--program--gene factorization framework. At the input, SpaFactor efficiently fuses the visual representation of the central spot with multiscale local and regional neighborhood context, yielding a histologic representation that captures cellular morphology and microenvironmental heterogeneity. For modeling, a residual MLP stably learns a nonlinear mapping from the tissue microenvironment to low-dimensional latent gene programs. These activities are decoded through shared gene loadings into coordinated multi-gene expression predictions. Across five public cohorts, SpaFactor achieves the best aggregate performance, with particularly clear improvements for spatially variable genes, and more faithfully recovers biologically organized spatial patterns. These results demonstrate that lightweight joint modeling of tissue context and gene programs can improve both predictive accuracy and biological fidelity.
\end{abstract}

\section{Introduction}

Spatial transcriptomics (ST) measures gene expression together with the tissue coordinates of each measurement, allowing expression differences to be localized to specific anatomical regions, tumor compartments, and cellular neighborhoods \cite{stahl2016,rao2021}. This spatial correspondence enables the molecular states of distinct tissue regions to be characterized within the tumor microenvironment, providing a basis for biomarker discovery, patient stratification, and studies of treatment response. The broader adoption of ST, however, remains constrained by assay cost, complex tissue processing, sequencing burden, and platform-specific workflows. Routinely collected H\&E-stained whole-slide images (WSIs) provide a complementary source of information: histologic patterns such as nuclear morphology, glandular architecture, stromal remodeling, necrosis, and immune infiltration reflect the cellular composition and molecular activity of tissue. Learning the relationship between these morphologic patterns and gene expression could therefore enable spatial expression mapping in large retrospective pathology cohorts and in samples without matched ST measurements \cite{he2020,wang2025benchmark}.

Recent methods have advanced H\&E-to-ST inference along several complementary directions. Pathology foundation models strengthen spot-level morphology encoding; multi-resolution fusion, spatial transformers, and graph networks incorporate information across tissue regions; contrastive and generative approaches improve image--expression alignment \cite{xie2023,chung2024,xu2024,wang2025benchmark,weng2026}. Together, these developments show that accurate prediction benefits from both strong visual representations and spatial interaction. A related challenge lies in the gene output space. Gene expression co-varies through cell states, metabolic processes, translation, stress responses, and signaling programs, whereas conventional multi-output regression provides little explicit structure for these dependencies. GeneQuery highlighted gene--gene dependency as a key modeling issue \cite{xiong2024}; pathway- and expression-foundation approaches further demonstrate the value of biologically coordinated outputs \cite{majumder2026,fang2026}. These findings identify two complementary requirements for H\&E-to-ST inference: contextualizing morphology across tissue regions and structuring dependencies across gene outputs.

Meeting both requirements within a compact predictor remains challenging. Direct regression on frozen features offers an economical baseline with limited task-specific structuring. Spatial transformers, graph-attention networks, multi-branch architectures, and cross-modal generative systems provide richer interaction, accompanied by additional optimization and deployment costs \cite{wang2025benchmark,wang2025fmh2st,yin2026}. This trade-off motivates our central question: after fixing a strong pathology representation, can a lightweight downstream model connect spatial tissue evidence with coordinated gene programs?

SpaFactor accomplishes this through a lightweight morphology-to-program-to-gene workflow. It concatenates the frozen GigaPath embedding of the central spot with parameter-free local and regional within-slide summaries, uses a four-block residual MLP to infer latent program activities, and decodes the full gene panel through a shared loading matrix. This design expands the morphological field of view, stabilizes the nonlinear morphology-to-program mapping, and lets related genes share statistical strength. Once embeddings and neighborhood summaries are cached, training and inference use only batched dense operations. Tiered HVG weighting and a mild gene-wise correlation term prioritize spatially informative patterns; an enhanced variant, SpaFactor-Cal, adds an optional fold-safe training signal while retaining the same H\&E-only inference path.

Our contributions are threefold:

\begin{itemize}
\item \textbf{Lightweight spatial context modeling.} We introduce parameter-free, precomputable spatial aggregation and a four-block residual MLP that unifies central morphology with local and regional tissue context without graph message passing, cross-spot attention, or encoder fine-tuning.
\item \textbf{Factorized gene-program prediction.} We formulate a morphology-to-program-to-gene mapping with shared gene loadings, explicitly coupling related targets and providing structured regularization without external gene databases, pathway labels, or an expression foundation model.
\item \textbf{Predictive accuracy, computational efficiency, and biological validation.} Across five cohorts totaling 421 slides and 799,085 spots, leakage-controlled evaluation, matched foundation-feature controls, ablations, and efficiency analysis establish the model's accuracy--efficiency advantage; biomarker maps and held-out pathway activities show that the gains extend to biologically organized spatial function.
\end{itemize}

\section{Related Work}

\subsection{Histology-to-Expression Prediction}

Early H\&E-to-expression methods established patch- or slide-level regression. ST-Net predicted spot expression from local breast histology, while HE2RNA and hist2RNA mapped histological features to transcriptomic profiles at slide or patch resolution \cite{he2020,schmauch2020,mondol2023}. HisToGene introduced inter-spot transformers; Hist2ST and THItoGene combined convolution with transformer or graph relations \cite{pang2021,zeng2022,jia2023}. BLEEP and mclSTExp moved toward contrastive image--expression alignment, whereas GenST explored cross-modal latent generation \cite{xie2023,min2024,wood2026}. Pathology foundation encoders now provide substantially richer morphological representations \cite{xu2024}, but backbone choice, spatial fusion, and downstream predictors are often varied simultaneously across studies. SpaFactor fixes the pathology foundation representation to systematically examine which compact task structures remain necessary for accurate prediction.

\subsection{Spatial Context and Structured Gene Modeling}

A parallel line enriches spatial or biological structure. Adaptive spatial GNNs, TRIPLEX, DeepSpot, FmH2ST, HiFusion, and HESpotEx incorporate neighborhoods, multiple image scales, graph interactions, or regional context \cite{song2024,chung2024,nonchev2025,wang2025fmh2st,weng2026,yin2026}. iStar and GHIST extend inference to super-resolution or single-cell prediction \cite{zhang2024,fu2025}. On the output side, GeneQuery uses genes as semantic queries, DKAN imports gene knowledge for contrastive alignment, HINGE transfers gene dependencies from a single-cell foundation model, and PEaRL represents expression through pathway activity \cite{xiong2024,zhang2026,fang2026,majumder2026}. These studies establish that both tissue context and functional coherence matter, but they commonly require attention-heavy interaction, external gene semantics, pathway supervision, or an expression-side foundation model. SpaFactor offers a compact, factorized alternative: parameter-free spatial context structures the morphology input, while a data-learned factor space structures the gene output. A fully matched direct GigaPath-MLP baseline and exact-configuration ablations then isolate and validate the independent value of this dual structure.

\section{Method}

\begin{figure*}[!t]
\centering
\includegraphics[width=\textwidth]{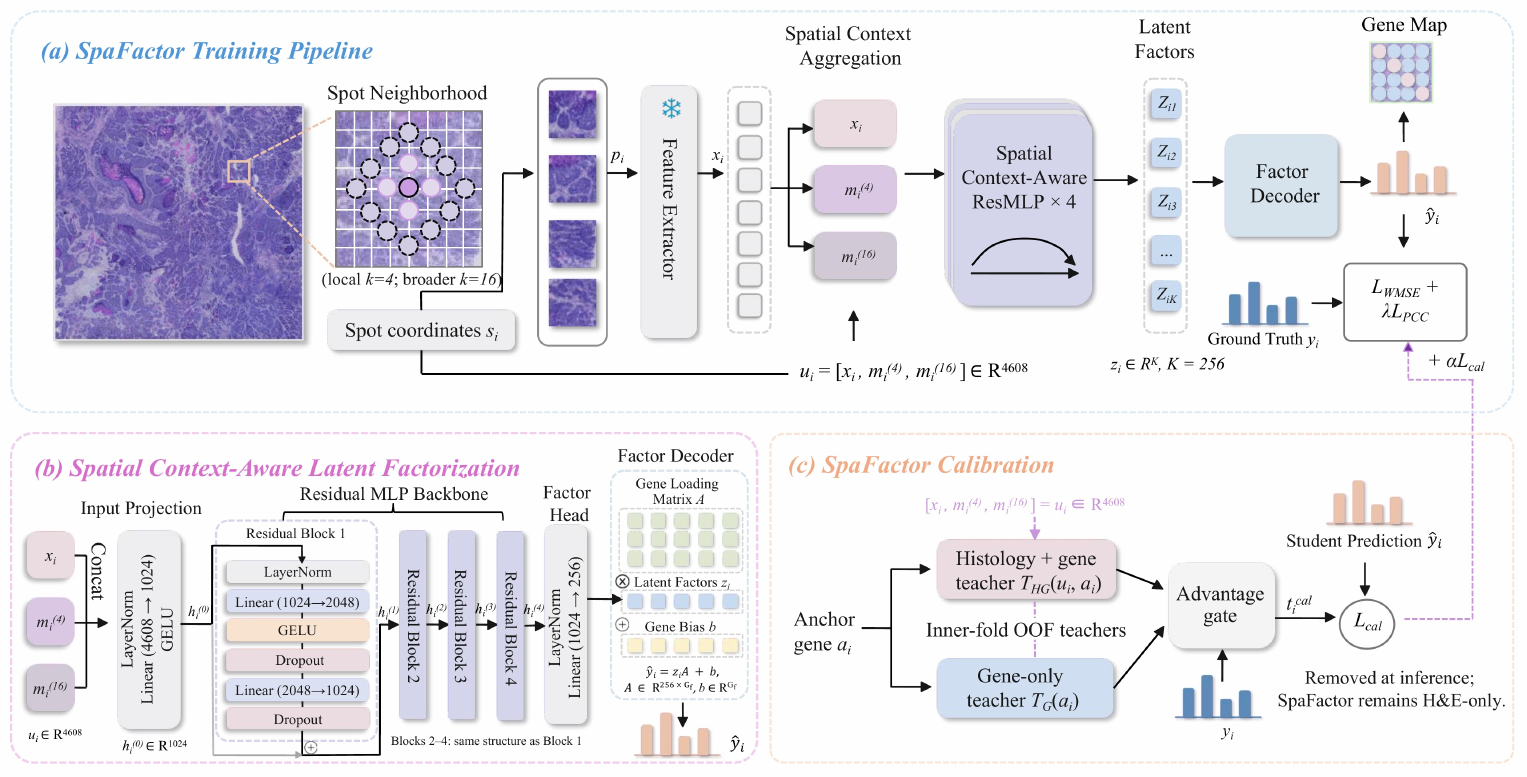}
\caption{Overview of SpaFactor. (a) The central spot's frozen GigaPath embedding is concatenated with mean embeddings from its local ($k=4$) and regional ($k=16$) neighborhoods, mapped by a four-block residual MLP to latent program activities, and decoded through shared gene loadings. (b) Detailed residual encoder and factor decoder. (c) Optional SpaFactor-Cal adds an inner-fold out-of-fold calibration loss during training and is removed for H\&E-only inference.}
\label{fig:1}
\end{figure*}

\subsection{Problem Formulation: H\&E-to-ST Prediction}

We formulate H\&E-to-ST inference as a spatially resolved multi-gene expression prediction task. Given paired histology images and spatial transcriptomic profiles during training, the model learns to predict the expression vector of a selected gene panel at each spot. At inference, only the H\&E image and spot coordinates are required.

Let spot $i$ have an H\&E crop $I_i$, coordinate $s_i$, slide identity $q_i$, and measured gene-count vector $c_i$. For outer fold $f$, the output panel $\mathcal{G}_f$ is constructed from training slides only. Counts are transformed by $\log(1+c)$, then standardized with training-fold gene means $\mu_g$ and standard deviations $\sigma_g$:

\begin{equation}
y_{ig} =
\frac{\log(1+c_{ig})-\mu_g}{\max(\sigma_g,10^{-3})}.
\end{equation}

SpaFactor learns $F_\theta:(I_i,s_i,q_i)\mapsto\hat{y}_i\in\mathbb{R}^{|\mathcal{G}_f|}$. The learning objective is to recover both spot-level expression profiles and gene-specific spatial variation. Predictions are inverse-standardized using the training-fold statistics and evaluated in log-expression space. Gene-panel construction, normalization, HVG ranking, and checkpoint selection are performed within the training fold.

\subsection{Frozen Histology Feature Extraction}

Each spot-centered H\&E crop is encoded once by the pretrained Prov-GigaPath tile encoder \cite{xu2024}, producing $x_i\in\mathbb{R}^{1536}$. The encoder remains frozen in every SpaFactor, SpaFactor-Cal, and matched GigaPath-MLP run. Precomputing these embeddings reduces training cost and prevents differences in visual-backbone fine-tuning from confounding the task-model comparison.

\subsection{Spatial Context Aggregation}

Coordinates define neighbors only among spots from the same slide:

\begin{equation}
\begin{aligned}
\mathcal{N}_k(i)=
\operatorname{kNN}\!\left(s_i;\{s_j:q_j=q_i,\ j\neq i\}\right),
\\
m_i^{(k)}=\frac{1}{|\mathcal{N}_k(i)|}
\sum_{j\in\mathcal{N}_k(i)}x_j.
\end{aligned}
\end{equation}

We use a local neighborhood $k=4$ and a regional neighborhood $k=16$. The resulting context-aware representation is

\begin{equation}
u_i=[x_i;m_i^{(4)};m_i^{(16)}]\in\mathbb{R}^{4608}.
\end{equation}

This operator supplies local and regional tissue evidence without transferring information across slides. It intentionally excludes central-minus-neighbor residual contrasts; validation selected the simpler mean-context design.

\subsection{Residual Morphology-to-Program Encoder}

The context vector is normalized and projected to a 1024-dimensional hidden state:

\begin{equation}
h_i^{(0)}=\operatorname{GELU}\!\left(W_0\operatorname{LN}(u_i)+b_0\right).
\end{equation}

Four residual MLP blocks refine the representation:

\begin{equation}
\begin{aligned}
r_i^{(\ell)}
&=\operatorname{Dropout}\!\left(
\operatorname{GELU}\!\left(
W_{\ell,1}\operatorname{LN}(h_i^{(\ell)})
\right)\right),\\
h_i^{(\ell+1)}
&=h_i^{(\ell)}+W_{\ell,2}r_i^{(\ell)},
\quad \ell=0,\ldots,3.
\end{aligned}
\end{equation}

The inner width is 2048 and dropout is 0.10. A factor head maps the final hidden state to $K=256$ activities:

\begin{equation}
z_i=W_z\operatorname{LN}(h_i^{(4)})+b_z,
\qquad z_i\in\mathbb{R}^{256}.
\end{equation}

\subsection{Factor Decoder}

A trainable loading matrix $A\in\mathbb{R}^{256\times|\mathcal{G}_f|}$ and gene bias $b$ decode the expression panel:

\begin{equation}
\hat{y}_i=z_iA+b.
\end{equation}

The factor decoder couples genes through shared spot-level activities, so related targets draw on a common predictive basis instead of being fitted only through gene-specific output weights. This shared program space regularizes gene-specific fluctuations unsupported by morphology and exposes a direct interface between prediction and functional gene sets. The direct-decoder ablation replaces $z_iA+b$ with one hidden-to-gene linear map.

The computational path is lightweight by construction. GigaPath embeddings are extracted once, neighborhood summaries are parameter-free and precomputable, and both ResMLP and factor decoding use dense tensor operations. Downstream training therefore requires neither dynamic graph construction nor communication among spots. For a 2,000-gene panel, the factorized output module is 62.1\% smaller than a direct 1024-to-gene output module; the complete trainable downstream predictor contains 22.30 million parameters.

\subsection{Variance-Aware Correlation-Aligned Learning}

HVG ranks are computed on outer-training data. Genes ranked in the top 50, 51--100, 101--200, and the remainder receive weights 4, 3, 2, and 1. The weighted reconstruction loss is

\begin{equation}
\mathcal{L}_{\mathrm{WMSE}}=
\frac{1}{|\mathcal{B}||\mathcal{G}_f|}
\sum_{i\in\mathcal{B}}\sum_{g\in\mathcal{G}_f}
w_g(\hat{y}_{ig}-y_{ig})^2.
\end{equation}

Let $\rho_g$ denote the Pearson correlation between predictions and targets for gene $g$ across a mini-batch. We add

\begin{equation}
\begin{aligned}
\mathcal{L}_{\mathrm{PCC}}
=1-\frac{\sum_g w_g\rho_g}{\sum_g w_g},
\\
\mathcal{L}_{\mathrm{SpaFactor}}
=\mathcal{L}_{\mathrm{WMSE}}+
0.1\mathcal{L}_{\mathrm{PCC}}.
\end{aligned}
\end{equation}

The correlation term is deliberately small. Its role is to align spatial variation without replacing pointwise reconstruction.

\subsection{Optional Training-Time Calibration}

SpaFactor-Cal adds expression-side information only on outer-training slides. A gene-only teacher and a histology-plus-gene teacher are trained in three inner folds, so every calibration target is out of fold with respect to its slide. An advantage gate retains targets for which the multimodal teacher improves over the gene-only teacher by a fixed margin. The student adds a gated calibration loss with weight $\alpha=0.07$.

At validation and test time, anchor genes, teachers, and the gate are removed. SpaFactor-Cal therefore has the same H\&E-only inference inputs and factor decoder as SpaFactor.

\section{Experiments}

\subsection{Experimental Setup}

\textbf{Datasets and protocol.} We evaluate five public cohorts from STimage-1K4M and HEST-1k \cite{chen2024,jaume2024}, totaling 421 sections and 799,085 spots. The primary cohort contains 108 breast sections from the original Spatial Transcriptomics platform (45,306 spots). External evaluation uses 87 Visium breast sections (163,931 spots), 73 bowel sections (205,642 spots), 91 brain sections (322,572 spots), and 62 skin sections (61,634 spots). Every cohort follows matched slide-level fivefold splits. Gene-panel construction, normalization, HVG ranking, and checkpoint selection are repeated inside each outer fold; validation slides select the checkpoint, and test slides are evaluated only after training.

\textbf{Preprocessing and baselines.} Spot coordinates are registered with the corresponding WSI. Spot-centered RGB tiles use the dataset-supplied crop radius for STimage or the aligned HEST tile archives; the native field of view is therefore source specific. For physical reference, the nominal capture-spot diameters are approximately 100~$\mu$m for original ST and 55~$\mu$m for Visium. Each tile is center-cropped to $224\times224$, normalized with ImageNet statistics, and encoded once by the frozen Prov-GigaPath tile encoder. Counts undergo $\log(1+c)$ transformation. Genes are filtered and up to 2,000 HVGs are selected using outer-training slides only; training-fold means and standard deviations are then applied to validation and test data. We compare with ST-Net, BLEEP, DeepSpot, GenST, HisToGene, HE2RNA, and hist2RNA, together with a matched GigaPath-MLP. All methods use the same frozen features, fold-specific gene panels, data splits, and evaluation code.

\textbf{Evaluation and implementation.} Spot PCC measures agreement between predicted and observed gene profiles within spots, whereas Gene PCC measures spatial agreement for each gene across spots. VR-PCC@$K$ averages Gene PCC over the $K\in\{50,100,200\}$ most variable held-out genes. Values are reported on a percentage scale as fivefold means and standard deviations; matched-control uncertainty is estimated by hierarchical paired bootstrap over datasets and folds. SpaFactor uses hidden width 1024, inner width 2048, four residual blocks, 256 factors, and dropout 0.10. AdamW is run with learning rate $10^{-3}$, weight decay $10^{-5}$, and batch size 1024 for at most 100 epochs, with 15-epoch early stopping selected by validation VR-PCC@200. Experiments use Ubuntu 22.04.2, Python 3.10.20, PyTorch 2.13.0, and NVIDIA RTX 5090 GPUs. Each training process uses one GPU, while independent folds and ablations are parallelized across devices. Frozen embeddings and neighborhood summaries are cached and reused across runs.

\subsection{Main Results}

Table~\ref{tab:1} summarizes the five-cohort comparison. On the primary benchmark, SpaFactor outperforms the strongest published comparator on all five metrics, with the clearest margins in gene-wise and variance-ranked recovery, including 36.36\% Gene PCC and 63.41\% VR-PCC@200. SpaFactor-Cal remains close to the teacher-free model, indicating that optional calibration is complementary rather than necessary for the overall gain.

\begin{table*}[t]
\centering
\small
\setlength{\tabcolsep}{2pt}
\begin{tabular*}{\textwidth}{@{\extracolsep{\fill}}llrrrrr@{}}
\toprule
Dataset & Method & Spot PCC (\%) & Gene PCC (\%) & VR-PCC@50 (\%) & VR-PCC@100 (\%) & VR-PCC@200 (\%) \\
\midrule
Primary breast & Strongest prior (HE2RNA) & \valsd{71.19}{3.37} & \valsd{33.86}{4.10} & \valsd{68.03}{6.26} & \valsd{64.12}{6.45} & \valsd{59.49}{6.69} \\
 & GigaPath-MLP & \valsd{70.29}{3.60} & \valsd{33.10}{4.48} & \valsd{67.41}{6.61} & \valsd{63.44}{6.76} & \valsd{58.84}{7.01} \\
 & SpaFactor-Cal & \valsd{\textbf{71.73}}{3.28} & \valsd{36.33}{3.71} & \valsd{71.69}{6.17} & \valsd{67.79}{6.00} & \valsd{63.20}{6.23} \\
 & \textbf{SpaFactor} & \valsd{71.67}{3.29} & \valsd{\textbf{36.36}}{4.17} & \valsd{\textbf{72.05}}{6.55} & \valsd{\textbf{68.09}}{6.42} & \valsd{\textbf{63.41}}{6.68} \\
\midrule
Visium breast & Strongest prior (hist2RNA) & \valsd{\textbf{70.07}}{2.38} & \valsd{25.27}{5.69} & \valsd{40.98}{8.51} & \valsd{42.74}{7.68} & \valsd{44.55}{8.99} \\
 & GigaPath-MLP & \valsd{68.54}{2.10} & \valsd{24.39}{4.34} & \valsd{38.62}{6.80} & \valsd{40.63}{6.09} & \valsd{42.54}{7.14} \\
 & SpaFactor-Cal & \valsd{69.39}{2.59} & \valsd{25.53}{5.37} & \valsd{\textbf{41.91}}{6.94} & \valsd{\textbf{43.61}}{6.84} & \valsd{\textbf{45.16}}{8.18} \\
 & \textbf{SpaFactor} & \valsd{69.25}{2.66} & \valsd{\textbf{25.65}}{5.40} & \valsd{41.49}{7.65} & \valsd{43.08}{7.99} & \valsd{44.59}{9.20} \\
\midrule
Bowel & Strongest prior (hist2RNA) & \valsd{\textbf{57.40}}{4.57} & \valsd{48.78}{6.80} & \valsd{63.33}{7.20} & \valsd{61.85}{6.56} & \valsd{59.36}{6.44} \\
 & GigaPath-MLP & \valsd{55.93}{5.00} & \valsd{48.73}{11.50} & \valsd{63.57}{10.54} & \valsd{62.08}{10.44} & \valsd{59.69}{10.52} \\
 & SpaFactor-Cal & \valsd{57.35}{5.13} & \valsd{\textbf{51.20}}{9.33} & \valsd{\textbf{66.55}}{7.93} & \valsd{\textbf{64.69}}{8.00} & \valsd{\textbf{62.07}}{8.32} \\
 & \textbf{SpaFactor} & \valsd{57.18}{5.30} & \valsd{50.31}{8.97} & \valsd{66.11}{8.00} & \valsd{64.17}{8.03} & \valsd{61.34}{8.26} \\
\midrule
Brain & Strongest prior (HE2RNA) & \valsd{60.84}{2.45} & \valsd{30.16}{6.08} & \valsd{51.73}{8.76} & \valsd{48.30}{8.39} & \valsd{43.41}{8.15} \\
 & GigaPath-MLP & \valsd{60.51}{2.22} & \valsd{29.28}{6.63} & \valsd{51.09}{9.62} & \valsd{47.49}{9.24} & \valsd{42.54}{8.87} \\
 & SpaFactor-Cal & \valsd{\textbf{61.12}}{2.38} & \valsd{\textbf{32.84}}{5.44} & \valsd{55.38}{7.73} & \valsd{\textbf{51.42}}{7.81} & \valsd{\textbf{46.28}}{7.64} \\
 & \textbf{SpaFactor} & \valsd{61.07}{2.08} & \valsd{32.79}{5.51} & \valsd{\textbf{55.39}}{7.22} & \valsd{51.41}{7.41} & \valsd{46.25}{7.39} \\
\midrule
Skin & Strongest prior (hist2RNA) & \valsd{67.41}{6.24} & \valsd{52.86}{4.16} & \valsd{79.69}{5.38} & \valsd{79.69}{5.58} & \valsd{78.52}{5.20} \\
 & GigaPath-MLP & \valsd{66.39}{5.98} & \valsd{52.81}{5.49} & \valsd{79.83}{7.10} & \valsd{79.88}{7.22} & \valsd{78.78}{6.96} \\
 & SpaFactor-Cal & \valsd{67.24}{7.47} & \valsd{53.51}{4.82} & \valsd{81.90}{6.28} & \valsd{81.67}{6.48} & \valsd{80.45}{6.25} \\
 & \textbf{SpaFactor} & \valsd{\textbf{67.61}}{7.11} & \valsd{\textbf{53.68}}{4.53} & \valsd{\textbf{82.32}}{5.84} & \valsd{\textbf{82.06}}{6.09} & \valsd{\textbf{80.88}}{5.70} \\
\bottomrule
\end{tabular*}
\caption{Five-cohort comparison. Values are fivefold means with standard deviations in lower-right subscripts. For each dataset, ``strongest prior'' denotes the published baseline with the best mean rank across the five metrics, with Gene PCC breaking ties. Bold marks the highest displayed mean; complete per-model results are reported separately.}
\label{tab:1}
\end{table*}

\begin{figure*}[!t]
\centering
\includegraphics[width=\textwidth]{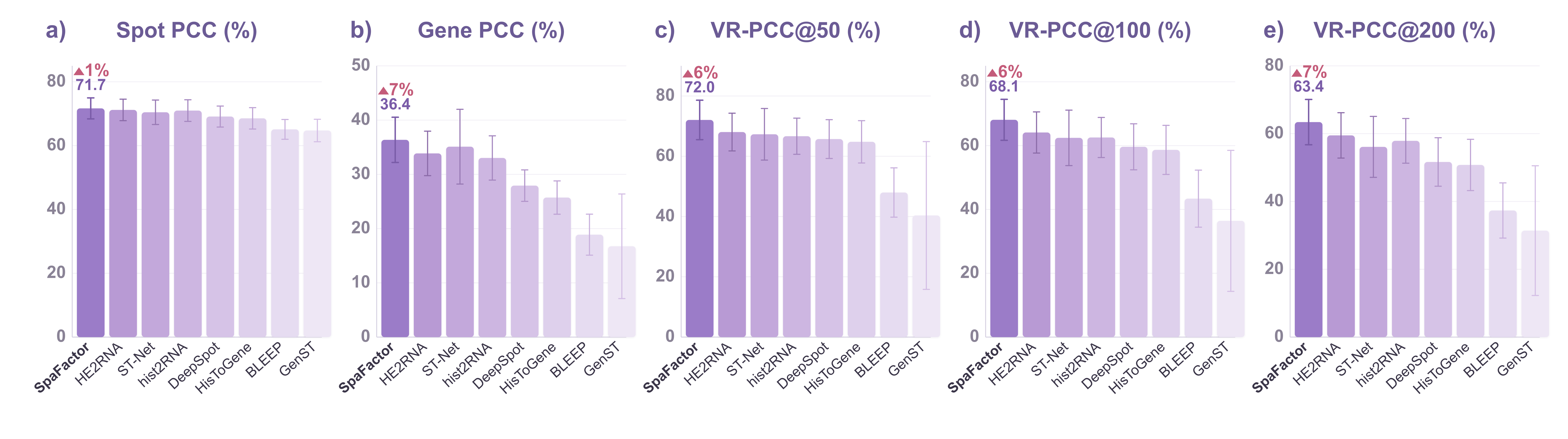}
\caption{Primary-cohort comparison with seven published methods. Panels show fivefold mean and standard deviation for Spot PCC, Gene PCC, and VR-PCC@50/100/200. Percentages show SpaFactor's relative gain over HE2RNA, the Table~\ref{tab:1} strongest prior for this cohort: $(\text{SpaFactor}-\text{HE2RNA})/\text{HE2RNA}\times100\%$.}
\label{fig:2}
\end{figure*}

Across the four external cohorts, SpaFactor or SpaFactor-Cal exceeds the selected strongest prior model on 18 of 20 displayed metrics. The advantage is most consistent for VR-PCC across all four tissues, showing that the framework primarily improves spatially organized expression rather than uniformly shifting every spot-level score. Figure~\ref{fig:2} confirms the same pattern in the complete primary-cohort comparison. The similar performance of SpaFactor and SpaFactor-Cal further suggests that most of the benefit comes from the core spatial-context and gene-program structure.

The metric profile clarifies the improvement. Spot PCC emphasizes agreement among genes within each spot, whereas Gene PCC and VR-PCC require each gene's spatial ordering to be recovered across tissue. Their larger, persistent gains from VR-PCC@50 through VR-PCC@200 therefore indicate broader spatial discrimination rather than an advantage driven by a few favorable markers.

\textbf{Matched Foundation-Feature Control.} Against the matched GigaPath-MLP, SpaFactor wins 20--23 of 25 dataset--fold comparisons per metric. Across these pairs, the relative gain---mean paired difference divided by the mean baseline---reaches 5.61\% for VR-PCC@50; every hierarchical paired-bootstrap interval excludes zero. Because the models share frozen embeddings, gene panels, folds, objective family, and evaluation code, this advantage isolates SpaFactor's downstream spatial-context and gene-program structure from the image encoder. The gains require neither graph propagation, cross-spot attention, nor end-to-end foundation-model tuning.

\subsection{Ablation and Sensitivity Analyses}

\subsubsection{Structural Component Ablations}

Removing both neighborhood summaries produces the largest interpretable decline in Table~\ref{tab:2}: Gene PCC falls by 2.37 percentage points and VR-PCC by about four percentage points. Regional context alone recovers most of the full model's performance, while local context adds a smaller improvement. The contextual gain therefore comes mainly from regional tissue organization, with the immediate neighborhood providing complementary detail.

\begin{table}[!t]
\centering
\small
\setlength{\tabcolsep}{2.2pt}
\renewcommand{\arraystretch}{1.10}
\resizebox{\columnwidth}{!}{%
\begin{tabular}{@{}l*{5}{c}@{}}
\toprule
\ablationheadfirst{Variant} & \ablationheadleft{Spot PCC} & \ablationheadleft{Gene PCC} & \ablationheadleft{VR-PCC@50} & \ablationheadleft{VR-PCC@100} & \ablationheadleft{VR-PCC@200} \\
\midrule
Full SpaFactor & \valsd{\textbf{71.67}}{3.29} & \valsd{\textbf{36.36}}{4.17} & \valsd{\textbf{72.05}}{6.55} & \valsd{\textbf{68.09}}{6.42} & \valsd{\textbf{63.41}}{6.68} \\
\midrule
\(\Delta\) Spot only & \valsd{70.59}{3.51} & \valsd{33.99}{3.96} & \valsd{68.07}{6.04} & \valsd{64.10}{6.22} & \valsd{59.61}{6.32} \\
\(\Delta\) Local only & \valsd{71.51}{3.46} & \valsd{35.66}{4.11} & \valsd{70.98}{6.56} & \valsd{67.00}{6.43} & \valsd{62.28}{6.72} \\
\(\Delta\) Regional only & \valsd{71.66}{3.36} & \valsd{36.14}{4.05} & \valsd{71.88}{6.25} & \valsd{67.87}{6.35} & \valsd{63.12}{6.59} \\
\(\Delta\) Plain MLP & \valsd{64.30}{1.52} & \valsd{16.74}{13.59} & \valsd{34.57}{24.55} & \valsd{32.86}{23.69} & \valsd{31.49}{23.15} \\
\(\Delta\) Direct decoder & \valsd{71.50}{3.30} & \valsd{35.74}{4.25} & \valsd{71.61}{7.10} & \valsd{67.66}{6.92} & \valsd{62.91}{7.09} \\
\(\Delta\) Uniform HVG weights & \valsd{71.51}{3.12} & \valsd{36.15}{4.08} & \valsd{71.88}{5.96} & \valsd{67.76}{6.17} & \valsd{63.05}{6.50} \\
\(\Delta\) No Gene-PCC loss & \valsd{71.59}{3.32} & \valsd{36.22}{4.19} & \valsd{71.95}{6.56} & \valsd{67.93}{6.41} & \valsd{63.27}{6.61} \\
\bottomrule
\end{tabular}
}
\caption{Structural component ablations on the primary benchmark. All metrics are percentages; values are fivefold means with standard deviations in lower-right subscripts.}
\label{tab:2}
\end{table}

Removing residual skips causes a much larger collapse, with VR-PCC@200 falling from 63.41\% to 31.49\%, which confirms that residual parameterization is necessary to optimize the deeper morphology-to-program mapping reliably. Direct gene decoding produces a smaller but consistent loss, whereas changing HVG weights or removing the Gene-PCC term has little effect. The ablations thus separate the roles of the components: spatial context supplies the main predictive signal, residual connections enable stable optimization of the deeper mapping, and factorized decoding provides a targeted refinement for coordinated genes.

\subsubsection{Sensitivity to Latent Factor Capacity}

Performance remains stable over $K=64$--1024 (Table~\ref{tab:3}). Although $K=512$ is marginally better on the primary cohort, its largest improvement over $K=256$ is only 0.33 percentage points, and increasing capacity to $K=1024$ brings no further gain.

\begin{table}[!t]
\centering
\small
\setlength{\tabcolsep}{2.2pt}
\renewcommand{\arraystretch}{1.10}
\resizebox{\columnwidth}{!}{%
\begin{tabular}{@{}l*{5}{c}@{}}
\toprule
\ablationheadfirst{$K$} & \ablationheadleft{Spot PCC} & \ablationheadleft{Gene PCC} & \ablationheadleft{VR-PCC@50} & \ablationheadleft{VR-PCC@100} & \ablationheadleft{VR-PCC@200} \\
\midrule
64 & \valsd{71.60}{3.56} & \valsd{36.18}{4.15} & \valsd{71.41}{6.25} & \valsd{67.45}{6.51} & \valsd{62.84}{6.70} \\
128 & \valsd{71.67}{3.50} & \valsd{35.82}{4.34} & \valsd{71.40}{6.54} & \valsd{67.31}{6.53} & \valsd{62.53}{6.80} \\
256 & \valsd{\textbf{71.67}}{3.29} & \valsd{36.36}{4.17} & \valsd{72.05}{6.55} & \valsd{68.09}{6.42} & \valsd{63.41}{6.68} \\
512 & \valsd{71.55}{3.32} & \valsd{\textbf{36.46}}{4.07} & \valsd{\textbf{72.38}}{6.14} & \valsd{\textbf{68.41}}{6.26} & \valsd{\textbf{63.72}}{6.56} \\
1024 & \valsd{71.57}{3.31} & \valsd{36.06}{4.29} & \valsd{72.02}{6.30} & \valsd{67.99}{6.46} & \valsd{63.24}{6.87} \\
\bottomrule
\end{tabular}
}
\caption{Sensitivity to latent factor count on the primary benchmark. All metrics are percentages; values are fivefold means with standard deviations in lower-right subscripts.}
\label{tab:3}
\end{table}

This small primary-cohort advantage does not generalize consistently to the external cohorts: $K=512$ helps consistently only on Visium breast and is weaker on most comparisons in bowel, brain, and skin. This pattern indicates that the model is not strongly capacity-limited and supports $K=256$ as the more transferable accuracy--capacity trade-off.

\subsubsection{Residual Depth and Efficiency}

\begin{figure}[!t]
\centering
\includegraphics[width=\columnwidth]{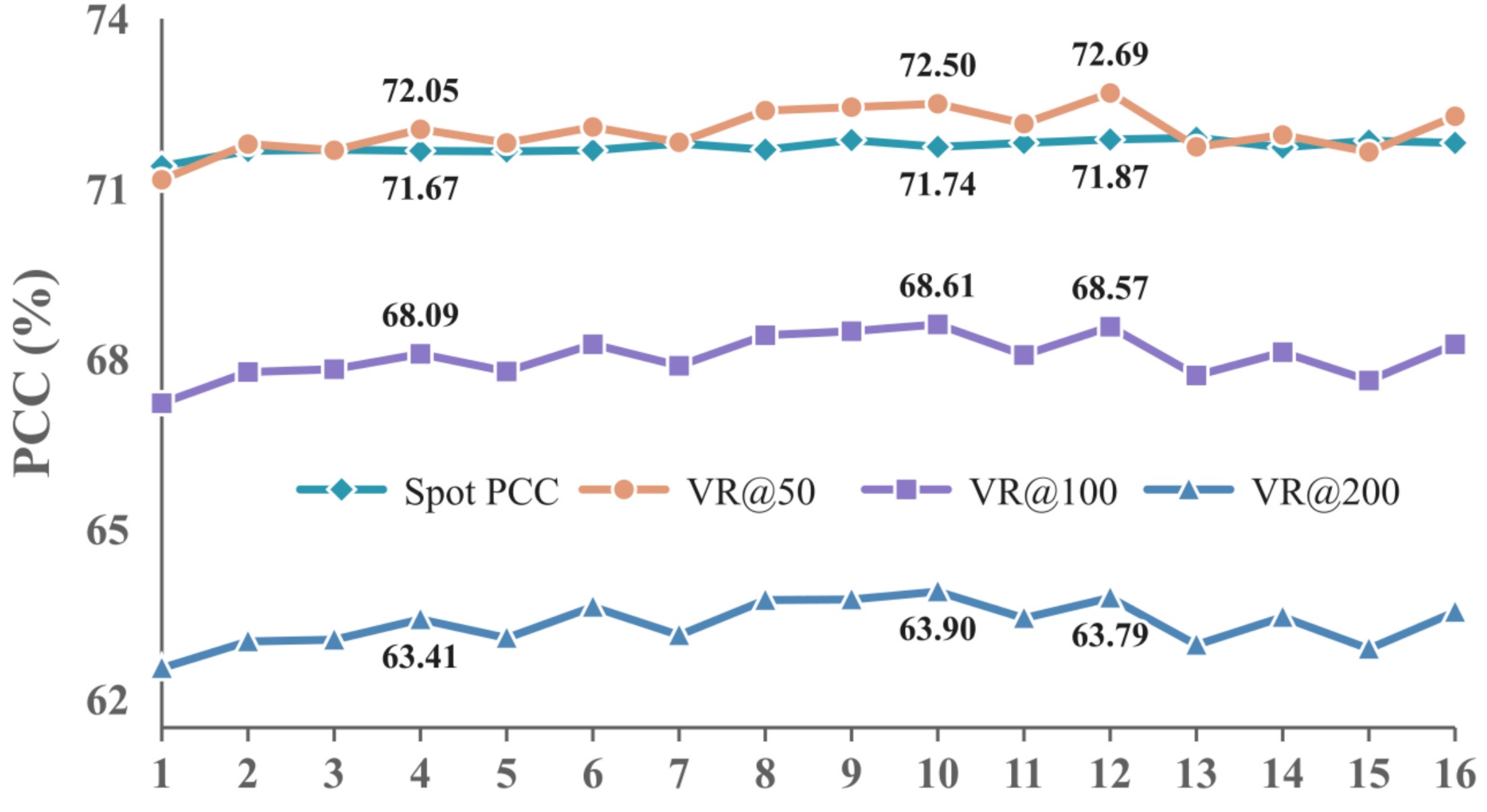}
\caption{Residual-depth sensitivity on the primary benchmark. Curves show fivefold mean Spot PCC (diamonds) and VR-PCC@50/100/200 (circles/squares/triangles); all metrics are percentages. Values are annotated at $L=4$, 10, and 12.}
\label{fig:depth}
\end{figure}

Residual depth has a non-monotonic effect (Figure~\ref{fig:depth}). One block is insufficient, while deeper models enter a broad high-performing region around $L=8$--12; beyond this range, no metric improves consistently. Depth therefore offers limited additional capacity rather than a uniform scaling benefit.

The modest accuracy changes come with a clear cost: $L=10$ and $L=12$ contain $2.13\times$ and $2.51\times$ as many trainable parameters as $L=4$, with up to 38\% longer training (Table~\ref{tab:4}). We therefore retain $L=4$ as the lightweight global configuration and treat the deeper settings as higher-capacity references.

\begin{table}[t]
\centering
\scriptsize
\setlength{\tabcolsep}{2pt}
\resizebox{\columnwidth}{!}{%
\begin{tabular}{lccc}
\toprule
\ablationheadleft{Depth} & \ablationheadleft{Trainable parameters} & \ablationheadleft{Training time (s)} & \ablationheadleft{Full-fold inference (s)} \\
\midrule
$L=4$ & 22,304,976 & 58.33 & 0.0315 \\
$L=10$ & 47,501,520 & 67.02 & 0.0365 \\
$L=12$ & 55,900,368 & 80.42 & 0.0398 \\
\bottomrule
\end{tabular}
}
\caption{Controlled depth--efficiency comparison on one held-out primary-cohort fold.}
\label{tab:4}
\end{table}

\subsubsection{Training Objective Sensitivity}

The model is also insensitive to the auxiliary Gene-PCC weight over $\lambda_{\mathrm{PCC}}=0$--0.20 (Table~\ref{tab:5}). The selected value of 0.10 is best overall, but improves the reported gene-level measures by at most 0.16 percentage points over removing the term. Together with the HVG-weight ablation, this shows that loss design refines performance but does not drive the main gain.

\begin{table}[!t]
\centering
\small
\setlength{\tabcolsep}{2.2pt}
\renewcommand{\arraystretch}{1.10}
\resizebox{\columnwidth}{!}{%
\begin{tabular}{@{}l*{5}{c}@{}}
\toprule
\ablationheadfirst{$\lambda_{\mathrm{PCC}}$} & \ablationheadleft{Spot PCC} & \ablationheadleft{Gene PCC} & \ablationheadleft{VR-PCC@50} & \ablationheadleft{VR-PCC@100} & \ablationheadleft{VR-PCC@200} \\
\midrule
0.00 & \valsd{71.59}{3.32} & \valsd{36.22}{4.19} & \valsd{71.95}{6.56} & \valsd{67.93}{6.41} & \valsd{63.27}{6.61} \\
0.05 & \valsd{\textbf{71.67}}{3.37} & \valsd{35.99}{4.08} & \valsd{71.70}{6.39} & \valsd{67.69}{6.34} & \valsd{62.97}{6.60} \\
0.10 & \valsd{\textbf{71.67}}{3.29} & \valsd{\textbf{36.36}}{4.17} & \valsd{\textbf{72.05}}{6.55} & \valsd{\textbf{68.09}}{6.42} & \valsd{\textbf{63.41}}{6.68} \\
0.20 & \valsd{71.57}{3.33} & \valsd{36.23}{4.10} & \valsd{71.81}{6.90} & \valsd{67.86}{6.64} & \valsd{63.18}{6.81} \\
\bottomrule
\end{tabular}
}
\caption{Gene-PCC loss-weight sensitivity on the primary benchmark. All metrics are percentages; values are fivefold means with standard deviations in lower-right subscripts. $\lambda_{\mathrm{PCC}}=0.10$ is the main setting.}
\label{tab:5}
\end{table}

Overall, the sensitivity analyses reveal a clear hierarchy: architecture matters more than capacity or loss tuning. The selected $K=256$, $L=4$ configuration favors transferability and computational economy over a small gain on a single cohort.

\begin{figure*}[!t]
\centering
\begin{minipage}{\textwidth}
\centering
\includegraphics[width=0.94\linewidth]{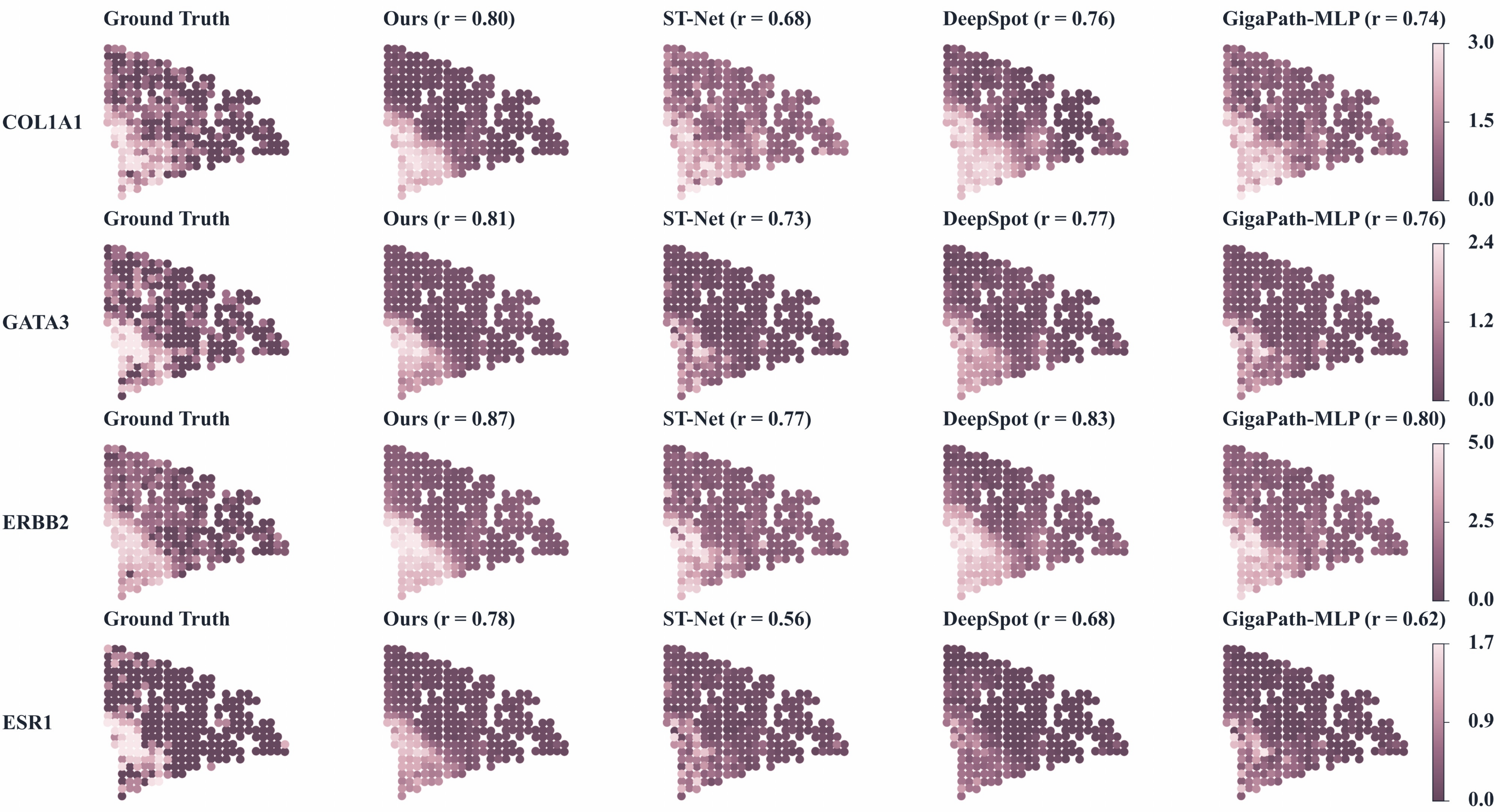}
\captionof{figure}{Spatial recovery of four breast biomarkers on one held-out section. Rows show COL1A1, GATA3, ERBB2, and ESR1; columns show measured expression and predictions from SpaFactor (``Ours''), ST-Net, DeepSpot, and matched GigaPath-MLP on the same spots. Pearson $r$ measures agreement with the measured map, and each row shares a common color scale.}
\label{fig:biomarkers}
\end{minipage}
\end{figure*}

\begin{figure}[!t]
\centering
\includegraphics[width=\columnwidth]{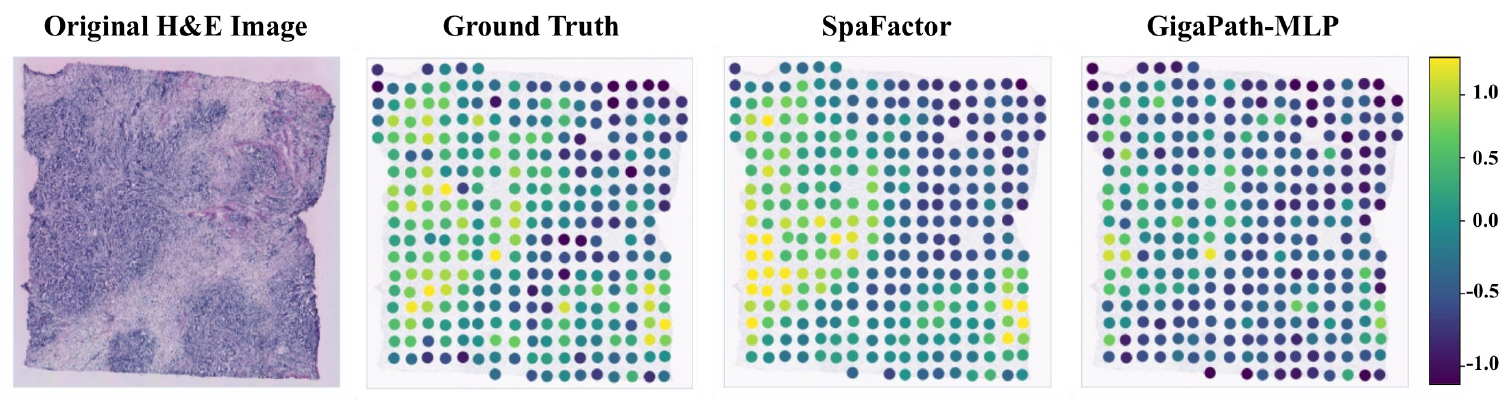}
\caption{Spatial recovery of KEGG oxidative-phosphorylation activity on a held-out section. The H\&E image, measured activity, and predictions from SpaFactor and matched GigaPath-MLP are aligned on the same 301 spots and share a $z$-score color scale.}
\label{fig:pathway-map}
\end{figure}

\begin{figure}[!t]
\centering
\includegraphics[width=\columnwidth]{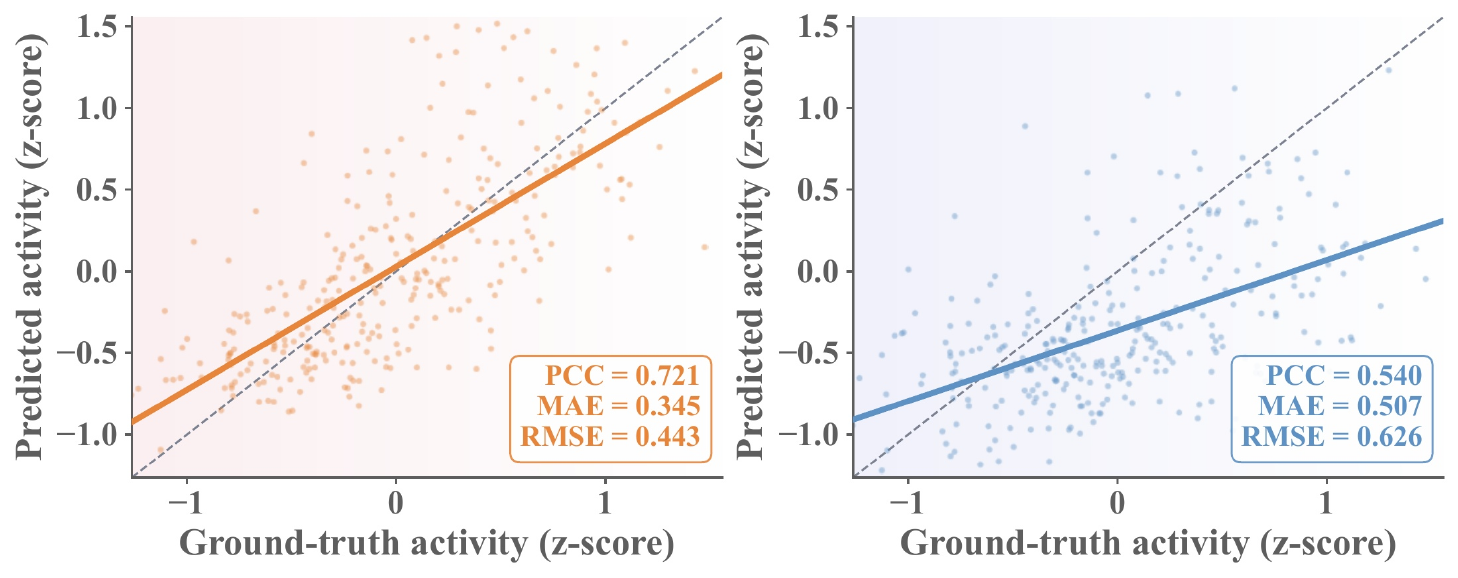}
\caption{Pathway-activity fit on the held-out section in Figure~\ref{fig:pathway-map}. Each point is one spot; solid lines are fitted regressions and dashed lines indicate identity. SpaFactor is shown on the left and matched GigaPath-MLP on the right; PCC, MAE, and RMSE are reported in each panel.}
\label{fig:pathway-fit}
\end{figure}

\FloatBarrier
\subsection{Biomarker and Functional Pathway Recovery}

Across the four breast markers in Figure~\ref{fig:biomarkers}, SpaFactor reaches PCCs of 0.78--0.87 and outperforms every displayed alternative. More importantly, it preserves the dominant expression territories and their boundaries across stromal, luminal, HER2, and estrogen-associated programs, whereas competing maps are more diffuse or attenuated. The aggregate gains therefore correspond to recognizable biological spatial organization.

These markers span extracellular-matrix-rich stroma (COL1A1) and distinct epithelial tumor programs (GATA3, ERBB2, and ESR1). Their simultaneous recovery argues against simple spatial smoothing and shows that the shared predictor retains program-specific compartments and boundaries.

Across 1,274 slide--pathway pairs from 108 held-out slides, SpaFactor improves pathway PCC by 6.11 percentage points on average (95\% bootstrap CI [4.71, 7.48]) and yields positive mean gains on 94 slides, extending its advantage from individual genes to coordinated biological function.

In the representative oxidative-phosphorylation example, SpaFactor preserves high-activity regions more faithfully, raises PCC from 0.540 to 0.721, and reduces MAE and RMSE (Figures~\ref{fig:pathway-map} and~\ref{fig:pathway-fit}). Together, these results show that gene-program modeling recovers both spatial ordering and activity magnitude without pathway-activity supervision.

\section{Conclusion}

SpaFactor demonstrates that a compact downstream model can connect spatial tissue context with coordinated gene programs on frozen pathology representations. Across five public cohorts, combining central morphology with local and regional context and decoding latent programs through shared gene loadings improves H\&E-to-ST prediction, especially for spatially variable genes. Matched controls and ablations identify spatial context as the primary performance source, while residual mapping and factorized decoding support stable, coordinated prediction. Biomarker and held-out pathway analyses further confirm biologically organized spatial recovery. Overall, SpaFactor provides an accurate, biologically faithful, and computationally efficient framework for H\&E-based spatial expression inference.

\bibliography{spafactor_references}
\clearpage

\input{supp2}

\end{document}

%% file: supp2.tex
\urlstyle{rm}
\def\UrlFont{\rm}
\frenchspacing

\pdfinfo{
/TemplateVersion (2027.1)
}

\setcounter{secnumdepth}{2}
\numberwithin{equation}{section}
\numberwithin{table}{section}
\numberwithin{figure}{section}

\newcommand{\ablationplaceholder}[1]{%
  \fbox{\parbox[c][0.13\textheight][c]{0.94\textwidth}{\centering\itshape #1}}%
}

\title{Supplementary Material for\\
SpaFactor: Lightweight Spatial Context Aware Gene Program Modeling for Histology to Transcriptomics Inference}
\author{Anonymous Authors}
\affiliations{Anonymous Institution}
\setcounter{dbltopnumber}{3}
\renewcommand{\dbltopfraction}{0.98}
\renewcommand{\dblfloatpagefraction}{0.90}

\maketitle
\appendix

\section{Detailed experimental protocol}

\subsection{Cohorts and evaluation splits}
\begin{table}[htp]
\centering
\scriptsize
\caption{Cohorts used in the anonymous submission.}
\resizebox{\columnwidth}{!}{%
\begin{tabular}{lccc}
\toprule
Cohort & Sections & Spots & Platform \\
\midrule
Primary breast & 108 & 45,306 & Spatial Transcriptomics \\
Visium breast & 87 & 163,931 & Visium \\
Bowel & 73 & 205,642 & Visium \\
Brain & 91 & 322,572 & Visium \\
Skin & 62 & 61,634 & Visium \\
\midrule
Total & 421 & 799,085 & N/A \\
\bottomrule
\end{tabular}
}
\label{tab:cohorts}
\end{table}

We evaluated five public spatial transcriptomics cohorts from STimage-1K4M and HEST-1k. The resulting cohort sizes are reported in Table~\ref{tab:cohorts}.

Each cohort used matched fivefold splits at the slide level (shuffle seed 42). Each outer fold held out one fifth of slides for testing; 12.5\% of the remainder was sampled for validation with seed $42+f$, yielding approximately 70\%/10\%/20\% train/validation/test partitions.

\subsection{Histology preprocessing}

Spot coordinates were registered with the corresponding WSI. For physical reference, the nominal capture spot diameters are approximately 100~$\mu$m for the original Spatial Transcriptomics platform and 55~$\mu$m for Visium. RGB tiles were cropped centrally to $224\times224$ pixels and normalized using ImageNet mean $(0.485,0.456,0.406)$ and standard deviation $(0.229,0.224,0.225)$. A frozen Prov-GigaPath tile encoder produced one 1,536-dimensional representation per spot. The same cached representation was used by SpaFactor, SpaFactor-Cal, and the matched GigaPath-MLP.

\begin{table}[htp]
\centering
\scriptsize
\caption{Final SpaFactor configuration without a teacher, used for the main results.}
\begin{tabular}{@{}p{0.26\columnwidth}p{0.67\columnwidth}@{}}
\toprule
Component & Setting \\
\midrule
Histology encoder & Frozen Prov-GigaPath, 1,536 dimensions \\
Context & Central spot + mean embeddings within the slide for $k=4$ and $k=16$ \\
Predictor & Residual MLP, hidden width 1,024, inner width 2,048, 4 blocks \\
Decoder & 256 factors with a shared gene loading matrix \\
Regularization & Dropout 0.10; AdamW weight decay $10^{-5}$ \\
Optimization & AdamW; learning rate $10^{-3}$; batch size 1,024; gradient clip 5.0; at most 100 epochs \\
Selection & Validation VR-PCC@200; patience 15; minimum improvement $10^{-5}$ \\
Loss & Tiered HVG weights $4/3/2/1$ and Gene-PCC weight 0.10 \\
Randomness & Split seed 42; primary result training seed 42 \\
\bottomrule
\end{tabular}
\label{tab:final-config}
\end{table}

\subsection{Expression preprocessing}

Prediction targets were $\log(1+c)$ values of raw counts. In each outer fold, candidate genes were constructed using expression from the training slides of that fold. A candidate had to be present in the gene schema of every training section, detected in at least 90\% of training sections, detected in at least 1\% of all training spots, and detected in at least 5\% of training spots within the cohort. After normalization to $10^4$ counts per library and $\log(1+\cdot)$ transformation, the lowest 5\% of genes by variance in normalized expression were removed.

HVG scores were then estimated using at most 2,000 randomly sampled spots from each training section. Genes were divided into 20 mean expression bins, a linear relation between log variance and log mean was fitted within each bin, and the standardized residual variance was converted to a percentile rank. The final gene panel contained at most 2,000 genes present in every section of the cohort. Its target composition was 1,200 consensus HVGs, 400 cohort specific HVGs, and 400 stable highly expressed genes; quotas were proportionally reduced when fewer eligible candidates were available. The values normalized by library size were used for filtering and ranking. Model targets and evaluation remained in the $\log(1+c)$ space of raw counts.

For each fold, the selected targets were standardized using the gene mean and standard deviation from the training fold, with the denominator bounded below by $10^{-3}$. These training statistics were applied unchanged to validation and test slides. Predictions were inverse standardized before evaluation. Thus, validation/test expression influenced neither gene filtering, panel ranking, target normalization, nor checkpoint selection.

\subsection{Evaluation metrics}

For a test set with $N$ spots and $G$ genes, let $r(\cdot,\cdot)$ denote Pearson correlation. We report
\[
\mathrm{Spot\ PCC}=\frac{1}{N}\sum_{i=1}^{N}r(\hat{\mathbf y}_{i:},\mathbf y_{i:}),
\]
\[
\mathrm{Gene\ PCC}=\frac{1}{G}\sum_{g=1}^{G}r(\hat{\mathbf y}_{:g},\mathbf y_{:g}),
\]
\[
\mathrm{VR\text{-}PCC@}K=\frac{1}{K}\sum_{g\in\mathcal V_K}r(\hat{\mathbf y}_{:g},\mathbf y_{:g}).
\]
where $\mathcal V_K$ contains the $K$ genes with the greatest expression variance in the held out data of the fold. VR-PCC therefore uses a variance ranking of the test fold for evaluation; it is distinct from the HVG ranking based on training data and used for the loss. All reported values are fivefold means and standard deviations on a percentage scale.

\subsection{Matched feature control}

The matched control uses direct gene regression from the same frozen GigaPath features; it shares folds, gene panels, loss weights, and evaluation code with SpaFactor. Table~\ref{tab:matched-control} summarizes all 25 pairs formed by datasets and folds. We use a hierarchical paired percentile bootstrap with 20,000 replicates, resampling datasets and then folds.

\begin{table}[htp]
\centering
\scriptsize
\caption{Final SpaFactor minus matched GigaPath-MLP across five cohorts and five folds.}
\resizebox{\columnwidth}{!}{%
\begin{tabular}{lccc}
\toprule
Metric & Mean delta (pp) & Wins & 95\% CI (pp) \\
\midrule
Spot PCC & +1.02 & 23/25 & [0.63, 1.45] \\
Gene PCC & +2.09 & 20/25 & [0.80, 3.44] \\
VR-PCC@50 & +3.37 & 21/25 & [2.07, 4.67] \\
VR-PCC@100 & +3.06 & 21/25 & [1.62, 4.44] \\
VR-PCC@200 & +2.82 & 21/25 & [1.27, 4.25] \\
\bottomrule
\end{tabular}
}
\label{tab:matched-control}
\end{table}

All five mean differences favored SpaFactor, with 95\% confidence intervals entirely above zero. The numbers of wins across the 25 paired comparisons by cohort and fold were 23, 20, 21, 21, and 21 for Spot PCC, Gene PCC, VR-PCC@50, VR-PCC@100, and VR-PCC@200, respectively. The largest gain was observed for VR-PCC@50 (+3.37 percentage points), showing that the context aware factorized predictor adds value beyond the shared frozen histology encoder.

\section{SpaFactor-Cal}

SpaFactor-Cal adds a calibration procedure during training. Anchor genes are selected from the training slides of the outer fold and are disjoint from the output panel. Three inner folds generate out of fold (OOF) targets, ensuring that each calibration prediction comes from a teacher not trained on that slide. We fit a gene based teacher $T_G(\mathbf a_i)$ and a teacher using histology and anchors, $T_{HG}(\mathbf u_i,\mathbf a_i)$, where $\mathbf u_i$ is the frozen tile embedding and $\mathbf a_i$ is the anchor expression.

For output gene $g$, the OOF multimodal advantage is
\[
\Delta_{ig}=\left[T_G(\mathbf a_i)_g-y_{ig}\right]^2-
\left[T_{HG}(\mathbf u_i,\mathbf a_i)_g-y_{ig}\right]^2.
\]
The gate $q_{ig}\in\{0,1\}$ retains entries whose advantage exceeds the fixed calibration margin. With the same tiered gene weight $w_g$ as the primary loss, the calibration term is
\[
\mathcal L_{\mathrm{cal}}=\frac{1}{BG}\sum_{i=1}^{B}\sum_{g=1}^{G}q_{ig}w_g\left(\hat y_{ig}-T_{HG}(\mathbf u_i,\mathbf a_i)_g\right)^2.
\]
SpaFactor-Cal optimizes $\mathcal L_{\mathrm{SpaFactor}}+0.07\mathcal L_{\mathrm{cal}}$. Teachers, anchors, and the gate are discarded before evaluation; inference therefore uses the same H\&E tiles and spatial coordinates as SpaFactor and requires no expression measurements.

\section{Reproducibility}
\subsection{Initialization stability}
With split seed 42 fixed, SpaFactor without a teacher was trained with seeds $1, 10, 42, 1234$, and $3407$ over the five folds of the primary cohort, using Table~\ref{tab:final-config} and zero calibration weight.

\begin{table*}[!t]
\centering
\small
\caption{Initialization stability across five training seeds; PCC mean $\pm$ standard deviation over five folds (\%).}
\begin{tabular*}{\textwidth}{@{\extracolsep{\fill}}lccccc@{}}
\toprule
Seed & Spot PCC (\%) & Gene PCC (\%) & VR-PCC@50 (\%) & VR-PCC@100 (\%) & VR-PCC@200 (\%) \\
\midrule
1    & \meanstd{71.52}{3.37} & \meanstd{36.51}{3.93} & \meanstd{72.08}{6.12} & \meanstd{68.24}{6.04} & \meanstd{63.60}{6.35} \\
10   & \meanstd{71.62}{3.27} & \meanstd{35.94}{3.75} & \meanstd{71.76}{5.78} & \meanstd{67.72}{5.88} & \meanstd{63.01}{5.99} \\
42   & \meanstd{71.67}{3.29} & \meanstd{36.36}{4.17} & \meanstd{72.05}{6.55} & \meanstd{68.09}{6.42} & \meanstd{63.41}{6.68} \\
1234 & \meanstd{71.65}{3.30} & \meanstd{36.06}{3.94} & \meanstd{71.61}{6.67} & \meanstd{67.60}{6.61} & \meanstd{62.86}{6.76} \\
3407 & \meanstd{71.66}{3.36} & \meanstd{35.93}{4.27} & \meanstd{71.35}{6.75} & \meanstd{67.33}{6.80} & \meanstd{62.60}{7.01} \\
\bottomrule
\end{tabular*}
\label{tab:seed-stability}
\end{table*}
Standard deviations across seeds were 0.06, 0.26, 0.31, 0.37, and 0.41 percentage points for Spot PCC, Gene PCC, and VR-PCC@50/100/200; the maximum range was 1.00 percentage points (VR-PCC@200).

\subsection{Compute environment}

Experiments used Ubuntu 22.04.2, Python 3.10.20, PyTorch 2.13.0/CUDA 13.0, NumPy 2.2.6, AMD EPYC 9655 CPUs, 377~GiB RAM, and 32~GB NVIDIA RTX 5090 GPUs. Each process used one GPU; cached tile embeddings and neighbourhood summaries were reused.

\section{Complete benchmark results}

Tables~\ref{tab:primary-all} through \ref{tab:skin-all} report the complete comparisons for the five formal cohorts. All methods were evaluated with the same folds, gene panels constructed from training data, frozen histology features, source normalization, and evaluation code.

\paragraph{Summary.}
Across five cohorts, the SpaFactor family led all 15 VR-PCC comparisons and achieved the highest Gene PCC in three cohorts; SpaFactor-Cal also led Spot PCC in two cohorts. This supports consistent gene-wise recovery with complementary calibration gains.

\begin{table*}[!t]
\centering
\small
\caption{Complete source normalized benchmark for the primary breast cohort. Values are fivefold mean $\pm$ standard deviation (\%); bold marks the highest mean in each column.}
\begin{tabular*}{\textwidth}{@{\extracolsep{\fill}}lccccc@{}}
\toprule
Method & Spot PCC (\%) & Gene PCC (\%) & VR-PCC@50 (\%) & VR-PCC@100 (\%) & VR-PCC@200 (\%) \\
\midrule
ST-Net & \meanstd{70.45}{3.83} & \meanstd{35.10}{6.89} & \meanstd{67.29}{8.56} & \meanstd{62.43}{8.69} & \meanstd{56.10}{8.99} \\
BLEEP & \meanstd{65.09}{3.10} & \meanstd{18.88}{3.78} & \meanstd{47.97}{8.18} & \meanstd{43.38}{8.94} & \meanstd{37.34}{8.14} \\
DeepSpot & \meanstd{69.11}{3.29} & \meanstd{27.92}{2.90} & \meanstd{65.70}{6.44} & \meanstd{59.63}{7.21} & \meanstd{51.65}{7.15} \\
GenST & \meanstd{64.74}{3.52} & \meanstd{16.74}{9.64} & \meanstd{40.35}{24.52} & \meanstd{36.42}{22.09} & \meanstd{31.40}{19.13} \\
HisToGene & \meanstd{68.55}{3.35} & \meanstd{25.72}{3.07} & \meanstd{64.79}{7.02} & \meanstd{58.66}{7.69} & \meanstd{50.77}{7.56} \\
HE2RNA & \meanstd{71.19}{3.37} & \meanstd{33.86}{4.10} & \meanstd{68.03}{6.26} & \meanstd{64.12}{6.45} & \meanstd{59.49}{6.69} \\
hist2RNA & \meanstd{70.98}{3.39} & \meanstd{33.02}{3.39} & \meanstd{66.65}{6.00} & \meanstd{62.55}{6.08} & \meanstd{57.88}{6.59} \\
GigaPath-MLP & \meanstd{70.29}{3.60} & \meanstd{33.10}{4.48} & \meanstd{67.41}{6.61} & \meanstd{63.44}{6.76} & \meanstd{58.84}{7.01} \\
SpaFactor-Cal & \textbf{\meanstd{71.73}{3.28}} & \meanstd{36.33}{3.71} & \meanstd{71.69}{6.17} & \meanstd{67.79}{6.00} & \meanstd{63.20}{6.23} \\
SpaFactor & \meanstd{71.67}{3.29} & \textbf{\meanstd{36.36}{4.17}} & \textbf{\meanstd{72.05}{6.55}} & \textbf{\meanstd{68.09}{6.42}} & \textbf{\meanstd{63.41}{6.68}} \\
\bottomrule
\end{tabular*}
\label{tab:primary-all}
\end{table*}

\begin{table*}[!t]
\centering
\small
\caption{Complete benchmark for the Visium breast cohort. Values are fivefold mean $\pm$ standard deviation (\%); bold marks the highest mean in each column.}
\begin{tabular*}{\textwidth}{@{\extracolsep{\fill}}lccccc@{}}
\toprule
Method & Spot PCC (\%) & Gene PCC (\%) & VR-PCC@50 (\%) & VR-PCC@100 (\%) & VR-PCC@200 (\%) \\
\midrule
ST-Net & \meanstd{69.85}{3.03} & \textbf{\meanstd{48.40}{6.84}} & \meanstd{38.91}{7.68} & \meanstd{39.28}{7.86} & \meanstd{39.95}{8.84} \\
BLEEP & \meanstd{68.37}{2.50} & \meanstd{13.58}{2.96} & \meanstd{26.61}{5.41} & \meanstd{26.10}{3.97} & \meanstd{24.13}{5.20} \\
DeepSpot & \textbf{\meanstd{70.46}{2.36}} & \meanstd{21.58}{4.25} & \meanstd{37.21}{4.06} & \meanstd{35.95}{4.53} & \meanstd{34.23}{5.70} \\
GenST & \meanstd{70.21}{2.46} & \meanstd{19.43}{3.02} & \meanstd{31.82}{4.58} & \meanstd{31.21}{4.78} & \meanstd{30.20}{5.09} \\
HisToGene & \meanstd{69.07}{2.42} & \meanstd{18.93}{4.17} & \meanstd{37.41}{3.84} & \meanstd{35.43}{4.85} & \meanstd{33.86}{6.32} \\
HE2RNA & \meanstd{69.88}{2.48} & \meanstd{24.85}{4.56} & \meanstd{39.56}{7.01} & \meanstd{41.78}{6.00} & \meanstd{43.89}{7.19} \\
hist2RNA & \meanstd{70.07}{2.38} & \meanstd{25.27}{5.69} & \meanstd{40.98}{8.51} & \meanstd{42.74}{7.68} & \meanstd{44.55}{8.99} \\
GigaPath-MLP & \meanstd{68.54}{2.10} & \meanstd{24.39}{4.34} & \meanstd{38.62}{6.80} & \meanstd{40.63}{6.09} & \meanstd{42.54}{7.14} \\
SpaFactor-Cal & \meanstd{69.39}{2.59} & \meanstd{25.53}{5.37} & \textbf{\meanstd{41.91}{6.94}} & \textbf{\meanstd{43.61}{6.84}} & \textbf{\meanstd{45.16}{8.18}} \\
SpaFactor & \meanstd{69.25}{2.66} & \meanstd{25.65}{5.40} & \meanstd{41.49}{7.65} & \meanstd{43.08}{7.99} & \meanstd{44.59}{9.20} \\
\bottomrule
\end{tabular*}
\label{tab:visium-all}
\end{table*}

\begin{table*}[!t]
\centering
\small
\caption{Complete bowel benchmark. Values are fivefold mean $\pm$ standard deviation (\%); bold marks the highest mean in each column.}
\begin{tabular*}{\textwidth}{@{\extracolsep{\fill}}lccccc@{}}
\toprule
Method & Spot PCC (\%) & Gene PCC (\%) & VR-PCC@50 (\%) & VR-PCC@100 (\%) & VR-PCC@200 (\%) \\
\midrule
ST-Net & \meanstd{53.59}{6.98} & \meanstd{36.87}{21.68} & \meanstd{47.88}{25.92} & \meanstd{46.42}{24.64} & \meanstd{43.38}{23.77} \\
BLEEP & \meanstd{50.71}{5.37} & \meanstd{15.76}{6.64} & \meanstd{30.87}{12.32} & \meanstd{27.67}{11.16} & \meanstd{23.82}{9.88} \\
DeepSpot & \meanstd{57.20}{5.67} & \meanstd{31.15}{6.56} & \meanstd{50.97}{8.97} & \meanstd{46.73}{8.34} & \meanstd{42.26}{7.78} \\
GenST & \meanstd{55.25}{5.52} & \meanstd{27.02}{6.26} & \meanstd{44.35}{8.81} & \meanstd{40.65}{8.35} & \meanstd{36.62}{7.81} \\
HisToGene & \meanstd{54.96}{5.50} & \meanstd{25.17}{7.00} & \meanstd{45.25}{10.12} & \meanstd{41.05}{9.76} & \meanstd{37.00}{8.89} \\
HE2RNA & \textbf{\meanstd{57.64}{5.20}} & \meanstd{43.78}{6.26} & \meanstd{60.20}{7.97} & \meanstd{58.46}{7.33} & \meanstd{55.76}{6.91} \\
hist2RNA & \meanstd{57.40}{4.57} & \meanstd{48.78}{6.80} & \meanstd{63.33}{7.20} & \meanstd{61.85}{6.56} & \meanstd{59.36}{6.44} \\
GigaPath-MLP & \meanstd{55.93}{5.00} & \meanstd{48.73}{11.50} & \meanstd{63.57}{10.54} & \meanstd{62.08}{10.44} & \meanstd{59.69}{10.52} \\
SpaFactor-Cal & \meanstd{57.35}{5.13} & \textbf{\meanstd{51.20}{9.33}} & \textbf{\meanstd{66.55}{7.93}} & \textbf{\meanstd{64.69}{8.00}} & \textbf{\meanstd{62.07}{8.32}} \\
SpaFactor & \meanstd{57.18}{5.30} & \meanstd{50.31}{8.97} & \meanstd{66.11}{8.00} & \meanstd{64.17}{8.03} & \meanstd{61.34}{8.26} \\
\bottomrule
\end{tabular*}
\label{tab:bowel-all}
\end{table*}

\begin{table*}[!t]
\centering
\small
\caption{Complete brain benchmark. Values are fivefold mean $\pm$ standard deviation (\%); bold marks the highest mean in each column.}
\begin{tabular*}{\textwidth}{@{\extracolsep{\fill}}lccccc@{}}
\toprule
Method & Spot PCC (\%) & Gene PCC (\%) & VR-PCC@50 (\%) & VR-PCC@100 (\%) & VR-PCC@200 (\%) \\
\midrule
ST-Net & \meanstd{60.07}{2.40} & \meanstd{27.66}{26.19} & \meanstd{40.18}{20.19} & \meanstd{36.71}{19.52} & \meanstd{32.23}{18.82} \\
BLEEP & \meanstd{55.73}{1.59} & \meanstd{16.28}{3.68} & \meanstd{34.10}{11.77} & \meanstd{29.89}{9.48} & \meanstd{24.31}{8.48} \\
DeepSpot & \meanstd{57.94}{1.79} & \meanstd{22.07}{2.29} & \meanstd{41.74}{8.78} & \meanstd{36.96}{6.63} & \meanstd{31.18}{5.51} \\
GenST & \meanstd{57.58}{2.22} & \meanstd{20.84}{3.02} & \meanstd{39.48}{9.65} & \meanstd{35.19}{7.63} & \meanstd{29.56}{6.90} \\
HisToGene & \meanstd{56.93}{2.42} & \meanstd{18.47}{3.62} & \meanstd{38.31}{11.85} & \meanstd{33.95}{9.24} & \meanstd{28.33}{8.02} \\
HE2RNA & \meanstd{60.84}{2.45} & \meanstd{30.16}{6.08} & \meanstd{51.73}{8.76} & \meanstd{48.30}{8.39} & \meanstd{43.41}{8.15} \\
hist2RNA & \meanstd{60.86}{2.02} & \meanstd{28.68}{7.46} & \meanstd{50.71}{9.94} & \meanstd{47.05}{9.77} & \meanstd{41.97}{9.63} \\
GigaPath-MLP & \meanstd{60.51}{2.22} & \meanstd{29.28}{6.63} & \meanstd{51.09}{9.62} & \meanstd{47.49}{9.24} & \meanstd{42.54}{8.87} \\
SpaFactor-Cal & \textbf{\meanstd{61.12}{2.38}} & \textbf{\meanstd{32.84}{5.44}} & \meanstd{55.38}{7.73} & \textbf{\meanstd{51.42}{7.81}} & \textbf{\meanstd{46.28}{7.64}} \\
SpaFactor & \meanstd{61.07}{2.08} & \meanstd{32.79}{5.51} & \textbf{\meanstd{55.39}{7.22}} & \meanstd{51.41}{7.41} & \meanstd{46.25}{7.39} \\
\bottomrule
\end{tabular*}
\label{tab:brain-all}
\end{table*}

\begin{table*}[!t]
\centering
\small
\caption{Complete skin benchmark. Values are fivefold mean $\pm$ standard deviation (\%); bold marks the highest mean in each column.}
\begin{tabular*}{\textwidth}{@{\extracolsep{\fill}}lccccc@{}}
\toprule
Method & Spot PCC (\%) & Gene PCC (\%) & VR-PCC@50 (\%) & VR-PCC@100 (\%) & VR-PCC@200 (\%) \\
\midrule
ST-Net & \meanstd{66.36}{6.76} & \textbf{\meanstd{63.04}{6.16}} & \meanstd{74.32}{9.72} & \meanstd{74.22}{10.01} & \meanstd{71.55}{9.96} \\
BLEEP & \meanstd{62.70}{6.44} & \meanstd{13.73}{5.54} & \meanstd{39.23}{12.02} & \meanstd{36.41}{8.84} & \meanstd{33.32}{9.16} \\
DeepSpot & \meanstd{65.32}{6.28} & \meanstd{24.76}{1.96} & \meanstd{52.59}{7.44} & \meanstd{49.92}{6.35} & \meanstd{47.98}{4.63} \\
GenST & \meanstd{64.49}{5.74} & \meanstd{21.36}{2.91} & \meanstd{46.45}{10.13} & \meanstd{45.11}{7.95} & \meanstd{43.33}{6.32} \\
HisToGene & \meanstd{64.93}{6.03} & \meanstd{20.73}{1.96} & \meanstd{52.59}{8.13} & \meanstd{49.76}{7.01} & \meanstd{47.45}{5.32} \\
HE2RNA & \textbf{\meanstd{68.05}{6.31}} & \meanstd{51.08}{4.78} & \meanstd{79.19}{6.09} & \meanstd{79.21}{6.28} & \meanstd{78.13}{5.99} \\
hist2RNA & \meanstd{67.41}{6.24} & \meanstd{52.86}{4.16} & \meanstd{79.69}{5.38} & \meanstd{79.69}{5.58} & \meanstd{78.52}{5.20} \\
GigaPath-MLP & \meanstd{66.39}{5.98} & \meanstd{52.81}{5.49} & \meanstd{79.83}{7.10} & \meanstd{79.88}{7.22} & \meanstd{78.78}{6.96} \\
SpaFactor-Cal & \meanstd{67.24}{7.47} & \meanstd{53.51}{4.82} & \meanstd{81.90}{6.28} & \meanstd{81.67}{6.48} & \meanstd{80.45}{6.25} \\
SpaFactor & \meanstd{67.61}{7.11} & \meanstd{53.68}{4.53} & \textbf{\meanstd{82.32}{5.84}} & \textbf{\meanstd{82.06}{6.09}} & \textbf{\meanstd{80.88}{5.70}} \\
\bottomrule
\end{tabular*}
\label{tab:skin-all}
\end{table*}